\pdfoutput=1

\documentclass[11pt]{article}

\usepackage{tabularx}
\usepackage[table]{xcolor}

\usepackage[]{acl}

\usepackage{times}
\usepackage{latexsym}

\usepackage[T1]{fontenc}

\usepackage[utf8]{inputenc}

\usepackage{microtype}

\usepackage{verbatim}
\usepackage{amsmath,graphicx}
\usepackage[ruled,vlined,linesnumbered]{algorithm2e}
\usepackage{tikz}
\usepackage[all]{hypcap}
\usepackage{anyfontsize}

\usepackage{kotex}

\usepackage[utf8]{inputenc}
\usepackage{dcolumn} 
\newcolumntype{d}[1]{D..{#1}}
\newcolumntype{P}[1]{>{\centering\arraybackslash}p{#1}}

\usepackage{booktabs,caption}
\usepackage[flushleft]{threeparttable} 

\newcommand\mc[1]{\multicolumn{1}{c}{#1}} 

\usepackage{multicol}
\usepackage{tabularray}
\usepackage{hyperref}

\title{KILDST: Effective Knowledge-Integrated Learning for Dialogue State Tracking using Gazetteer and Speaker Information}

\author{Hyungtak Choi\textsuperscript{1}, Hyeonmok Ko\textsuperscript{1}, Gurpreet Kaur\textsuperscript{2}, Lohith Ravuru\textsuperscript{1}, \\{\bf Kiranmayi Gandikota\textsuperscript{2}, Manisha Jhawar\textsuperscript{2}, Simma Dharani\textsuperscript{2}, and Pranamya Patil} \textsuperscript{2} \\ \textsuperscript{1}Samsung Research, Seoul, South Korea  \\ \textsuperscript{2}Samsung Research India Bangalore, Bangalore, India \\ 
\texttt{\{ht777.choi,felix.ko,k.gurpreet,loki.ravuru,k.gandikota} \\ 
\texttt{{m.jhawar,s.dharani,pran.patil}\}@samsung.com} \\}

\begin{document}

\maketitle
\begin{abstract}
Dialogue State Tracking (DST) is core research in dialogue systems and has received much attention. In addition, it is necessary to define a new problem that can deal with dialogue between users as a step toward the conversational AI that extracts and recommends information from the dialogue between users. So, we introduce a new task – DST from dialogue between users about scheduling an event (DST-USERS). The DST-USERS task is much more challenging since it requires the model to understand and track dialogue states in the dialogue between users and to understand who suggested the schedule and who agreed to the proposed schedule. To facilitate DST-USERS research, we develop dialogue datasets between users that plan a schedule. The annotated slot values which need to be extracted in the dialogue are date, time, and location. Previous approaches, such as Machine Reading Comprehension (MRC) and traditional DST techniques, have not achieved good results in our extensive evaluations. By adopting the knowledge-integrated learning method, we achieve exceptional results. The proposed model architecture combines gazetteer features and speaker information efficiently. Our evaluations of the dialogue datasets between users that plan a schedule show that our model outperforms the baseline model.
\end{abstract}

\section{Introduction}

\label{sec:intro}
DST is a task to determine the final dialogue states by continuously tracking the dialogue between the user and the system. It is a challenging and essential task because it can be applied to many real-world applications, such as voice assistant systems. Many approaches have been proposed to solve the DST problem \cite{lin2020mintl, somdst}.
MinTL \cite{lin2020mintl} framework adopts plug-and-play architecture to the pre-trained Seq2Seq model and can learn DST and NLU at the same time. SOM-DST \cite{somdst} updates the dialogue state in two steps: state operation prediction such as ADD, UPDATE and DELETE operation and dialogue state updater. The approaches have the advantage of improving performance; however, the error gets propagated to the dialogue state tracking phase of the model if an error occurs in the prediction of state operation. Meanwhile, in the past few years, many innovative models in the field of MRC.
Among them, the mainstream approach formalizes reading comprehension to the extent of extracting answers from a given text \cite{seo2016bidirectional, wang2016machine, xiong2017dcn+, joshi2017triviaqa, dunn2017searchqa, shen2017reasonet, wang2017evidence, wang2017gated, tan2017s, devlin2018bert, liu2019roberta}. There have been various attempts to apply this promising MRC technique to the DST field, and it has shown remarkable performance \cite{alexa2020a, alexa2020b}.

\begin{figure*}[tb]
	\centering
	\makebox[\textwidth]{\includegraphics[width=.9\paperwidth]{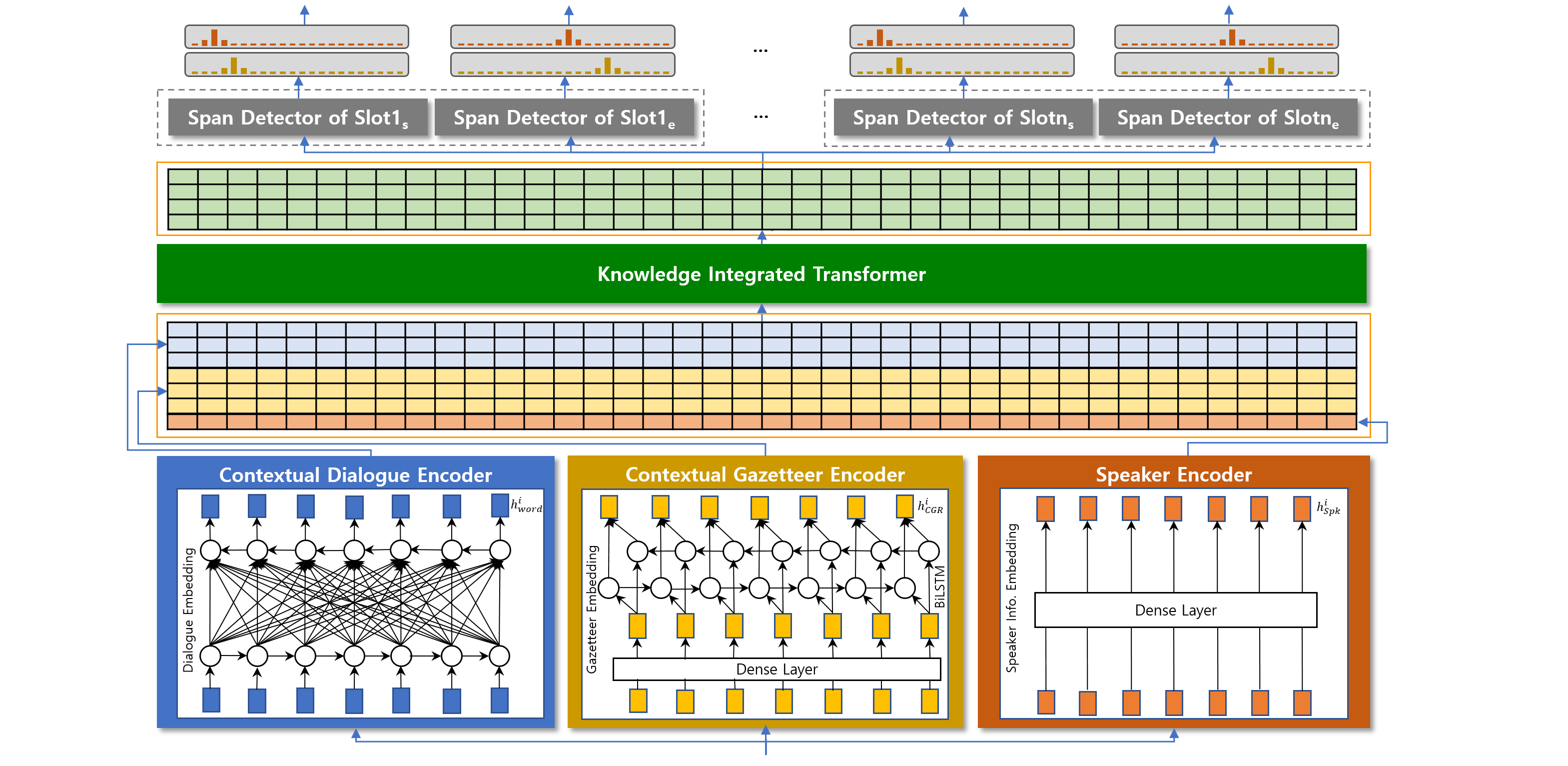}}
	\caption{Illustration of our proposed KILDST architecture. The model consists of 3 different sub-encoders, 1) contextual dialogue encoder for encoding dialogue, 2) contextual gazetteer encoder for encoding domain knowledge, and 3) speaker encoder for encoding speaker information. Our proposed knowledge-integrated learning method efficiently consolidates information from multiple encoders and extracts the schedule using span detectors.}
	\label{fig:model}
\end{figure*}

Despite much research on DST, there are still some challenging problems to be solved. Newly-coined words and unseen words pose problems \cite{bernier2020hardeval}. Some slots, such as specific store names and movie names, are not general noun phrases and are more complicated to recognize \cite{ashwini2014targetable, jayarao2018exploring}. To overcome this problem, approaches for integrating external knowledge, such as gazetteer information or knowledge based on neural architectures, have been highlighted and studied again. One-hot vectors are typically used as inputs to the gazetteer encoder and are then concatenated with word representations. However, for some problems, simply integrating gazetteer information will not improve or reduce performance for some slots \cite{meng2021gemnet}.

In this paper, we conduct a new and challenging task that tracks the dialogue state between two users rather than the dialogue state tracking in the dialogue between users and systems, like in previous studies. Ultimately, this model can be applied to a recommendation service on a smartphone that extracts and recommends a schedule for WhatsApp, WeChat, Kakaotalk, and Telegram.
Even if ambiguous slots are mentioned in the dialogue, the system cannot request explicit confirmation from the user, so it is more difficult to extract accurate information. 
Several studies use a gazetteer to improve performance \cite{song2020improving} to solve this problem. Especially, GEMNET \cite{meng2021gemnet} further enhances performance with effective gated gazetteer representations. We propose a novel architecture that effectively learns the gazetteer and speaker information to extract slots more accurately.

\noindent
The contributions of our work are as follows:
\begin{itemize}
	\vspace{-0.3cm}
	\item We propose a novel model based on the Transformer that efficiently consolidates the gazetteer knowledge and speaker information to improve the extraction performance of difficult words such as newly coined words and abbreviations used in dialogue between users.
	\vspace{-0.3cm}
	\item We propose a method of applying the GEMNET that efficiently utilizes the gazetteer knowledge in our integrated Transformer.
	\vspace{-0.3cm}
	\item We propose a method of understanding speaker information through speaker embeddings to know who suggested the schedule and who agreed.
	\vspace{-0.3cm}
	\item We evaluate our model for new dialogue datasets that contains a dialogue between users regarding scheduling an event and show that efficient use of gazetteer knowledge and speaker information improves performance.
\end{itemize}


\section{Proposed Model}
\label{sec:methods}
We propose an effective Knowledge-Integrated Learning method for Dialogue State Tracking using Gazetteer and Speaker Information (KILDST).
As shown in Figure 1, the proposed model consists of 3 different sub-encoders, 1) a contextual dialogue encoder for encoding dialogue, 2) a contextual gazetteer encoder for encoding domain knowledge, and 3) a speaker encoder for encoding speaker information. Our proposed knowledge-integrated learning method efficiently consolidates information from multiple encoders and extracts the slots related to a schedule, such as a date, time, and location using span detectors.

\subsection{Contextual Dialogue Encoder}
\label{sec:dsr}

A contextual dialogue encoder encodes the input dialogue based on a pre-trained BERT \cite{devlin2018bert}.
We use syllable units for tokens because this outperforms other token units in our experiments using Korean datasets.

\vspace{-0.4cm}
\begin{equation}			
\begin{array}{cl}
		h_{word} = BERT(D)
		\label{bert}
\end{array}
\end{equation}

\vspace{-0.1cm}
In the above formula, D refers to an index list represented by the dialogue text tokenized in syllable units in the form of "$[CLS]$ user A's dialogue sentence $[TURN]$ user B's dialogue sentence $[TURN]$". "$[TURN]$" is a special token for distinguishing the dialogue turn between users.

\subsection{Contextual Gazetteer Encoder}
\label{sec:dsr}

Gazetteer information can be provided directly as an input feature, but more is needed and sparse. We use linear projection to obtain a dense representation that captures interactions between multiple matches per syllable unit. We encode Contextual Gazetteer Representations (CGR) with gazetteer information. BiLSTM is then applied to contextualize this representation \cite{meng2021gemnet}.

\vspace{-0.5cm}
\begin{equation}		
\begin{array}{cl}
		h_{forward}^i = LSTM(h_{forward}^{i-1}, h_{gaz}^i)\\
		h_{backward}^i = LSTM(h_{backward}^{i+1}, h_{gaz}^i)\\
		h_{cgr}^i = [h_{forward}^i, h_{backward}^i]	
\end{array}
\end{equation}		

\subsection{Speaker Encoder}
\label{sec:dsr}

Our task differs from general dialogue system tasks between the user and the system. Since the dialogue is between two users and not between a user and a system, it is crucial to learn each dialogue information so that user A and user B can be distinguished. Therefore, it performs better when the model is accompanied by additionally providing a one-hot encoder to represent the speaker id for a given utterance.



\subsection{Knowledge-Integrated Transformer}
\label{sec:dsr}

We integrate contextual dialogue embedding, contextual gazetteer embedding, and speaker embedding. Especially, We integrate the Mixture of Experts (MoE) mechanism \cite{pavlitskaya2020using, meng2021gemnet} at the knowledge-integrated Transformer to utilize both the dialogue and gazetteer information efficiently.
We add gating networks to create a weighted linear combination of words and gazetteer representations. Training the gating network prevents the overuse or underuse of features.

\vspace{-0.7cm}
\begin{equation}				
	\begin{array}{cl}
		w_e = \sigma(\theta[h_{word}, h_{cgr}]),\\
		h_{moe} = w_e $·$ h_{word} + (1 - w_e) $·$ h_{cgr}
	\end{array}
	\label{GEMNET}
\end{equation}

\vspace{-0.7cm}
\begin{equation}
\begin{small}	
\begin{array}{cl}
		h_{integrated} = Transformer([h_{moe}, h_{spk}])
		\label{integrated Transformer Encoder}
\end{array}
\end{small}
\end{equation}	

\vspace{-0.1cm}
$h_{word}$, $h_{cgr}$ and $h_{spk}$ are the output of respective sub-modules. They are used to train the gating network. $\sigma$ is a Sigmoid activation function and $\theta$ is a trainable parameter. [.,.] represents a concatenation. We learn gating weights $w_e$.
The model can learn how to dynamically calculate each syllable unit's hidden information $h_{integrated}$. After obtaining $h_{integrated}$, we feed it to the integrated Transformer encoder to learn the integrated embedding vector.

\subsection{Span Detector}
\label{sec:dsr}

The span detector model, which uses a composite embedding vector, predicts the position information of all the slots related to the scheduled event. For all the slots of the scheduled events, the span detector is considered to take as the token level representation $[t_1, $· · ·$,t_n]$, the output of the knowledge-integrated Transformer. Each token representation $t_i$ is projected linearly through a common layer whose output values correspond to start and end positions. Softmax is then applied to the position values to produce a probability distribution for all tokens. Finally, we extract the span value with the highest probability distributions for each target slot and provide that as the output. The formula of the learning method of this model is as follows.

\vspace{-0.2cm}
\begin{equation}				
	\begin{array}{cl}
		P_s = W_s $·$ h_{integrated} + b_s	\\	
		P_e = W_e $·$ h_{integrated} + b_e	\\		
		P_{joint} = [P_s, P_e] \\				
		L_s = CCE(P_s, y_s)	\\		
		L_e = CCE(P_e, y_e)	\\		
		L_{joint} = JE(P_{joint}, y_{joint}) \\				
		L = L_s + L_e + L_{joint}
	\end{array}
	\label{model loss}
\end{equation}

\begin{center}
\begin{table*}[ht]
{\small
\hfill{}
\begin{tabular}{c}
\centering 
\begin{threeparttable}

\begin{tabular}{@{} l *{10}{d{3}} } 
\toprule

& \multicolumn{3}{c}{3K dataset} & \multicolumn{3}{c}{10K dataset} \\ 
\cmidrule(lr){2-4} \cmidrule(l){5-7}
\multicolumn{1}{c}{Model Type} & \mc{JGA} & \mc{Slot Acc.} & \mc{Slot F1} 
& \mc{JGA} & \mc{Slot Acc.} & \multicolumn{1}{c}{Slot F1}\\ 
\midrule
SOM-DST \cite{somdst}* & \mc{48.00} & \mc{-} & \mc{83.73} & \mc{62.40} & \mc{-} & \mc{89.33} \\ 
DSTRC (baseline) \cite{alexa2020a}* & \mc{50.70} & \mc{91.56} & \mc{84.97} & \mc{68.80} & \mc{94.47} & \mc{90.52} \\ 
BERT-SpanDetector & \mc{60.00} & \mc{94.12} & \mc{90.38} & \mc{73.00} & \mc{95.71} & \mc{92.47} \\ 
BERT-SpanDetector +Gaz. & \mc{63.67} & \mc{94.41} & \mc{91.18} & \mc{73.30} & \mc{95.60} & \mc{92.29} \\ 
BERT-SpanDetector +Gaz. +Transformer & \mc{66.67} & \mc{95.29} & \mc{92.26} & \mc{73.60} & \mc{95.80} & \mc{92.46} \\ 
BERT-SpanDetector +Gaz. +Transformer +Spk & \mc{68.00} & \mc{95.12} & \mc{91.84} & \mc{73.37} & \mc{95.80} & \mc{94.41} \\ 
BERT-SpanDetector +Context Gaz. +Transformer & \mc{68.33} & \mc{95.67} & \mc{92.29} & \mc{73.80} & \mc{95.78} & \mc{92.64} \\ 
BERT-SpanDetector +Context Gaz. +Transformer +MoE & \mc{69.44} & \mc{95.33} & \mc{92.29} & \mc{75.40} & \mc{96.10} & \mc{93.09} \\ 
BERT-SpanDetector +Context Gaz. +Transformer +MoE & \mc{\textbf{70.16}} & \mc{\textbf{95.35}} & \mc{\textbf{92.57}} & \mc{\textbf{77.80}} & \mc{\textbf{96.37}} & \mc{\textbf{93.61}} \\ 
\hspace{8.2em} +Spk \textbf{(KILDST)} & & & & & & \\ 
\midrule

\end{tabular} 

\end{threeparttable}
\end{tabular}}
\hfill{}
\caption{Results on the 3K and 10K datasets for all models . MoE is a mixture of experts. Spk is speaker embedding.}
\end{table*}
\end{center}

\vspace{-0.5cm}
In Equation (4) above, $h_{integrated}$ means integrated token on the knowledge-integrated Transformer, $W_s$ and $W_e$ mean the weight matrix, and $b_s$ and $b_e$ mean the bias. $P_s$ and $P_e$ mean the probability distribution of start and end positions for each token in the dialogue input. $y_s$ and $y_e$ denote the position labels for the correct answer range. Along with modeling start position and end position probabilities separately using Categorical Cross Entropy (CCE) loss, we use Jaccard Expectation (JE) loss to optimize start and end positions jointly instead of CCE loss. Because it showed better results when using JE loss.


\section{Experiments and Results}

\subsection{Datasets and Experimental Setup}

One of the essential goals of our work is to collect and create dialogue datasets between users that plan specific appointments. We collected and utilized their actual dialogue datasets in Korean with the consent of users to use the provided datasets for research. Also, to collect dialogue data of various ages, gender, and relationship combinations between users who have dialogue, we use a crowdsourcing platform. We created a dataset by providing chat rooms where real users can chat under certain predefined conditions. We provided the purpose of the dialogue, the relationship between users, and guidance information to these user chat rooms. If we set a specific profile of the chat room we want to collect, only users who meet the conditions can attend the chat room. We can collect various combinations and types of dialogue datasets by introducing a real-chat simulator between users in a constrained environment. We evaluate our models on two datasets, such as the 3K and 10K datasets. The 3K dataset has 3,000 dialogues, and 33,585 dialogue turns. 10K dataset has 10,000 dialogues, and 109,971 dialogue turns. The dataset was experimented with and evaluated by the ratio of train set 9 and test set 1. All models are implemented using Tensorflow 2.5.

\

\subsection{Evaluation Metrics and Results}

We use joint goal accuracy, slot accuracy, and slot F1 score to evaluate our model. \textbf{Joint Goal Accuracy} is an accuracy that checks whether all slot values predicted at a dialogue exactly match the ground truth values. The comparison of the results of our model with the state-of-the-art model using MRC techniques on the datasets is presented in Table 1. Our KILDST model achieves higher scores in all evaluation metrics in the test dataset. It shows that the joint goal accuracy increases in our experiments. Whenever we add new features such as speaker encoder, contextual gazetteer encoder, and a mixture of experts gating modules \cite{meng2021gemnet}, the joint goal accuracy increases in our experiments. The composite output vector consists of speaker embedding, contextual gazetteer embedding, and dialogue text embedding. Our best model extracts most slots better than the state-of-the-art model and also all other models in the evaluation dataset.

\vspace{-0.1cm}
\begin{table}[ht]
\begin{small}
\centering
\begin{tabular}{ p{0.27\linewidth} p{0.17\linewidth} p{0.17\linewidth} p{0.17\linewidth}}
\midrule
Slot Type & JGA & Slot Acc. & Slot F1\\
\midrule
Overall Slots & \textbf{77.80} & \textbf{96.37} & \textbf{93.61} \\ 
YEAR & 99.10 & 99.10 & 99.10 \\ 
MONTH & 99.10 & 99.10 & 99.10 \\ 
WEEK & 98.60 & 98.60 & 98.60 \\ 
DATE & 93.40 & 93.40 & 93.40 \\ 
AMPM & 93.20 & 93.20 & 93.20 \\ 
HOUR & 95.70 & 95.70 & 95.70 \\ 
MINUTE & 99.40 & 99.40 & 99.40 \\ 
LOCATION & 92.50 & 92.50 & 92.50 \\
\midrule
\end{tabular}
\end{small}
\hfill{}
\hfill{}
\hfill{}
\hfill{}
\caption{The best model overall slots result on the 10K}
\end{table}%

\vspace{-0.1cm}
\begin{table}[ht]
\begin{small}
\begin{center}
\centering
\begin{tabular}{ p{0.4\linewidth} P{0.26\linewidth} P{0.14\linewidth} }
\midrule
Hyperparameter & Search Range & Optimal Value\\
\midrule
Batch size & [16,32] & 32 \\ 
Epochs & [50,75,100] & 100 \\
Learning rate & - & 0.0001 \\ 
Optimizer & [Adam,AdamW] & AdamW \\ 
Dropout rate & - & 0.1 \\ 
BERT Max length & - & 512 \\ 
BERT Hidden size  & - & 256 \\ 
BiLSTM Input word size & - & 512 \\ 
BiLSTM Hidden size & - & 256 \\
Integrated Transformer \#layers & [1,2] & 2 \\
\midrule
\end{tabular}
\end{center}
\end{small}
\hfill{}
\caption{Hyperparameters for KILDST model}
\end{table}%

\section{Ablation Analysis}

\subsection{Effect of Gazetteer}
\label{sec:dsr}
Integrating additional gazetteer information into the model improves the performance of all models. In particular, models trained with 3,000 datasets benefit from gazetteer knowledge more than models trained with 10,000 datasets. Experiments using a gazetteer in 3,000 datasets have improved JGA performance by 3.67 \%. In the case of fine-turned models with insufficient training data, using the gazetteer knowledge has a more significant effect on improving performance. On the other hand, when training the models with a large number of training datasets, such as the model trained through 10,000 datasets, it is interpreted that it gets a relatively small benefit because much gazetteer information is already included in the training datasets. In addition, the results of the model using the contextual gazetteer encoder by the situation encoded by BiLSTM have improved the JGA performance by 1.66 \%in the experiments tested with 3,000 data sets than the simple one-hot gazetteer embedding. The results indicate the high efficiency of the CGR, which clearly shows the effect of the integration of the gazetteer.

\subsection{Effect of Mixture of Experts}
\label{sec:dsr}
The KILDST model is trained and fused with the BERT dialogue encoder and contextual gazetteer encoder. To assess the MoE component's impact, we concatenate the output vector of the contextual dialogue encoder and contextual dialogue encoder without MoE. By applying MoE, there was a 1.11\% JGA performance improvement effect in the model experiment with 3,000 datasets and a 1.6\% JGA performance improvement effect in the model experiment with 10,000 datasets. Table 1 shows more improvement in JGA performance than other accuracies when applying MoE.

\subsection{Effect of Integrated Transformer}
\label{sec:dsr}

In the case of our existing BERT-SpanDetector model, the output vectors of the contextual gazetteer representation and the contextual dialogue representation are concatenated and transferred to the span detector without additional training. However, our proposed KILDST approach, which once again trains through the Transformer Encoder, effectively integrates two output vectors, generates an updated embedding vector, and transfers it to the span detector. Additional training using an integrated transformer has improved JGA performance by 3\% in model experiments using 3,000 datasets.

\section{Conclusions}
This paper presents a novel model architecture that effectively learns the gazetteer knowledge and speaker information for DST. Our model consists of 3 different sub-encoders a contextual dialogue encoder for encoding dialogue, a contextual gazetteer encoder for encoding domain knowledge, and a speaker encoder for encoding speaker information. The knowledge-integrated learning method we proposed outperforms other models by integrating the information of various encoders more efficiently and accurately extracting slots. In addition, we have collected and created new dialogue datasets between users that plan a schedule. Our evaluation of this dialogue dataset shows improvement over the state-of-the-art model by better extracting the schedule from a dialogue containing difficult words such as newly coined words and abbreviations. Our model can be applied to a messenger app such as WhatsApp, WeChat, Kakaotalk, and Telegram as a recommendation system, extracting a schedule from dialogue and recommending a schedule to the user. In the future, we plan to expand our proposed model architecture to a model that extracts another meaningful event from a dialogue among more users.

\section*{Acknowledgement}
The authors would like to thank Chanwoo Kim, Munjo Kim, Jonggu Kim, Siddhartha Mukherjee, Raghavendra Hanumantasetty Ramasetty, and Aniruddha Uttam Tammewar for their helpful suggestions and discussion during this project.



\bibliography{refs, strings}

\begin{thebibliography}{21}
\expandafter\ifx\csname natexlab\endcsname\relax\def\natexlab#1{#1}\fi

\bibitem[{Ashwini and Choi(2014)}]{ashwini2014targetable}
Sandeep Ashwini and Jinho~D Choi. 2014.
\newblock Targetable named entity recognition in social media.
\newblock \emph{arXiv preprint arXiv:1408.0782}.

\bibitem[{Bernier-Colborne and Langlais(2020)}]{bernier2020hardeval}
Gabriel Bernier-Colborne and Philippe Langlais. 2020.
\newblock Hardeval: Focusing on challenging tokens to assess robustness of ner.
\newblock In \emph{Proceedings of the 12th Language Resources and Evaluation
  Conference}, pages 1704--1711.

\bibitem[{Devlin et~al.(2018)Devlin, Chang, Lee, and
  Toutanova}]{devlin2018bert}
Jacob Devlin, Ming-Wei Chang, Kenton Lee, and Kristina Toutanova. 2018.
\newblock Bert: Pre-training of deep bidirectional transformers for language
  understanding.
\newblock \emph{arXiv preprint arXiv:1810.04805}.

\bibitem[{Dunn et~al.(2017)Dunn, Sagun, Higgins, Guney, Cirik, and
  Cho}]{dunn2017searchqa}
Matthew Dunn, Levent Sagun, Mike Higgins, V~Ugur Guney, Volkan Cirik, and
  Kyunghyun Cho. 2017.
\newblock Searchqa: A new q\&a dataset augmented with context from a search
  engine.
\newblock \emph{arXiv preprint arXiv:1704.05179}.

\bibitem[{Gao et~al.(2020)Gao, Agarwal, Chung, Jin, and
  Hakkani-Tur}]{alexa2020b}
Shuyang Gao, Sanchit Agarwal, Tagyoung Chung, Di~Jin, and Dilek Hakkani-Tur.
  2020.
\newblock From machine reading comprehension to dialogue state tracking:
  Bridging the gap.
\newblock In \emph{Proceedings of the 2nd Workshop on Natural Language
  Processing for Conversational AI}.

\bibitem[{Gao et~al.(2019)Gao, Sethi, Agarwal, Chung, and
  Hakkani-Tur}]{alexa2020a}
Shuyang Gao, Abhishek Sethi, Sanchit Agarwal, Tagyoung Chung, and Dilek
  Hakkani-Tur. 2019.
\newblock Dialog state tracking: A neural reading comprehension approach.
\newblock In \emph{Proceedings of Special Interest Group on Discourse and
  Dialogue}.

\bibitem[{Jayarao et~al.(2018)Jayarao, Jain, and
  Srivastava}]{jayarao2018exploring}
Pratik Jayarao, Chirag Jain, and Aman Srivastava. 2018.
\newblock Exploring the importance of context and embeddings in neural ner
  models for task-oriented dialogue systems.
\newblock \emph{arXiv preprint arXiv:1812.02370}.

\bibitem[{Joshi et~al.(2017)Joshi, Choi, Weld, and
  Zettlemoyer}]{joshi2017triviaqa}
Mandar Joshi, Eunsol Choi, Daniel~S Weld, and Luke Zettlemoyer. 2017.
\newblock Triviaqa: A large scale distantly supervised challenge dataset for
  reading comprehension.
\newblock \emph{arXiv preprint arXiv:1705.03551}.

\bibitem[{Kim et~al.(2019)Kim, Yang, Kim, and Lee}]{somdst}
Sungdong Kim, Sohee Yang, Gyuwan Kim, and Sang{-}Woo Lee. 2019.
\newblock \href {http://arxiv.org/abs/1911.03906} {Efficient dialogue state
  tracking by selectively overwriting memory}.
\newblock \emph{CoRR}, abs/1911.03906.

\bibitem[{Lin et~al.(2020)Lin, Madotto, Winata, and Fung}]{lin2020mintl}
Zhaojiang Lin, Andrea Madotto, Genta~Indra Winata, and Pascale Fung. 2020.
\newblock Mintl: Minimalist transfer learning for task-oriented dialogue
  systems.
\newblock \emph{arXiv preprint arXiv:2009.12005}.

\bibitem[{Liu et~al.(2019)Liu, Ott, Goyal, Du, Joshi, Chen, Levy, Lewis,
  Zettlemoyer, and Stoyanov}]{liu2019roberta}
Yinhan Liu, Myle Ott, Naman Goyal, Jingfei Du, Mandar Joshi, Danqi Chen, Omer
  Levy, Mike Lewis, Luke Zettlemoyer, and Veselin Stoyanov. 2019.
\newblock Roberta: A robustly optimized bert pretraining approach.
\newblock \emph{arXiv preprint arXiv:1907.11692}.

\bibitem[{Meng et~al.(2021)Meng, Fang, Rokhlenko, and Malmasi}]{meng2021gemnet}
Tao Meng, Anjie Fang, Oleg Rokhlenko, and Shervin Malmasi. 2021.
\newblock Gemnet: Effective gated gazetteer representations for recognizing
  complex entities in low-context input.
\newblock In \emph{Proceedings of the 2021 Conference of the North American
  Chapter of the Association for Computational Linguistics: Human Language
  Technologies}, pages 1499--1512.

\bibitem[{Pavlitskaya et~al.(2020)Pavlitskaya, Hubschneider, Weber, Moritz,
  Huger, Schlicht, and Zollner}]{pavlitskaya2020using}
Svetlana Pavlitskaya, Christian Hubschneider, Michael Weber, Ruby Moritz,
  Fabian Huger, Peter Schlicht, and Marius Zollner. 2020.
\newblock Using mixture of expert models to gain insights into semantic
  segmentation.
\newblock In \emph{Proceedings of the IEEE/CVF Conference on Computer Vision
  and Pattern Recognition Workshops}, pages 342--343.

\bibitem[{Seo et~al.(2016)Seo, Kembhavi, Farhadi, and
  Hajishirzi}]{seo2016bidirectional}
Minjoon Seo, Aniruddha Kembhavi, Ali Farhadi, and Hannaneh Hajishirzi. 2016.
\newblock Bidirectional attention flow for machine comprehension.
\newblock \emph{arXiv preprint arXiv:1611.01603}.

\bibitem[{Shen et~al.(2017)Shen, Huang, Gao, and Chen}]{shen2017reasonet}
Yelong Shen, Po-Sen Huang, Jianfeng Gao, and Weizhu Chen. 2017.
\newblock Reasonet: Learning to stop reading in machine comprehension.
\newblock In \emph{Proceedings of the 23rd ACM SIGKDD International Conference
  on Knowledge Discovery and Data Mining}, pages 1047--1055.

\bibitem[{Song et~al.(2020)Song, Lawrie, Finin, and
  Mayfield}]{song2020improving}
Chan~Hee Song, Dawn Lawrie, Tim Finin, and James Mayfield. 2020.
\newblock Improving neural named entity recognition with gazetteers.
\newblock \emph{arXiv preprint arXiv:2003.03072}.

\bibitem[{Tan et~al.(2017)Tan, Wei, Yang, Du, Lv, and Zhou}]{tan2017s}
Chuanqi Tan, Furu Wei, Nan Yang, Bowen Du, Weifeng Lv, and Ming Zhou. 2017.
\newblock S-net: From answer extraction to answer generation for machine
  reading comprehension.
\newblock \emph{arXiv preprint arXiv:1706.04815}.

\bibitem[{Wang and Jiang(2016)}]{wang2016machine}
Shuohang Wang and Jing Jiang. 2016.
\newblock Machine comprehension using match-lstm and answer pointer.
\newblock \emph{arXiv preprint arXiv:1608.07905}.

\bibitem[{Wang et~al.(2017{\natexlab{a}})Wang, Yu, Jiang, Zhang, Guo, Chang,
  Wang, Klinger, Tesauro, and Campbell}]{wang2017evidence}
Shuohang Wang, Mo~Yu, Jing Jiang, Wei Zhang, Xiaoxiao Guo, Shiyu Chang, Zhiguo
  Wang, Tim Klinger, Gerald Tesauro, and Murray Campbell. 2017{\natexlab{a}}.
\newblock Evidence aggregation for answer re-ranking in open-domain question
  answering.
\newblock \emph{arXiv preprint arXiv:1711.05116}.

\bibitem[{Wang et~al.(2017{\natexlab{b}})Wang, Yang, Wei, Chang, and
  Zhou}]{wang2017gated}
Wenhui Wang, Nan Yang, Furu Wei, Baobao Chang, and Ming Zhou.
  2017{\natexlab{b}}.
\newblock Gated self-matching networks for reading comprehension and question
  answering.
\newblock In \emph{Proceedings of the 55th Annual Meeting of the Association
  for Computational Linguistics (Volume 1: Long Papers)}, pages 189--198.

\bibitem[{Xiong et~al.(2017)Xiong, Zhong, and Socher}]{xiong2017dcn+}
Caiming Xiong, Victor Zhong, and Richard Socher. 2017.
\newblock Dcn+: Mixed objective and deep residual coattention for question
  answering.
\newblock \emph{arXiv preprint arXiv:1711.00106}.

\end{thebibliography}

\newpage

\appendix
\section{Appendices}
\label{sec:appendix}

\begin{table}[ht]
\begin{small}
\begin{center}
\begin{tabularx}{\textwidth}[t]{XX}
\arrayrulecolor{black}\hline
\\
\textbf{\textcolor{black}{Korean conversation between 2 users}} & \textbf{\textcolor{black}{Korean conversation translated into English}}
\\
\\
\hline
\\
\hspace{0.2em}
\begin{minipage}[t]{\linewidth}
\begin{itemize}
\item[User1:] 예슬아 언제 시간됨?
\item[User2:] 왜?
\item[User1:] 용진이한테 빌린 보드게임 돌려줘야하는데 그런김에 같이 모이게
\item[User2:] ㅇㅋ 나 이번주는 안되고 다음주 일요일에 될듯? 저번처럼 스타벅스? 
\item[User1:] ㅇㅇ넴 저녁먹게 6시에 보죠
\item[User2:] 그때 다른 약속있거든... 그래서 한시간 늦게보자 \\
\item[User1:] 난 괜찮음
\end{itemize} 
\end{minipage}
\hspace{2em}
&
\hspace{1em}
\begin{minipage}[t]{\linewidth}
\begin{itemize}
\item[User1:] Jennie, When are you free?
\item[User2:] Why?
\item[User1:] I have to return the board game I borrowed from Yongjin, so we might as well meet up sometime
\item[User2:] Okay, I am not free this week, but next Sunday is good. Starbucks like last time?
\item[User1:] Let's meet up at 6 o'clock and have dinner
\item[User2:] I have an appointment at that time, but let's make it an hour late
\item[User1:] Sounds good\\
\end{itemize} 
\end{minipage}\\
\arrayrulecolor{black}\hline
\end{tabularx}
\end{center}
\end{small}
\hfill{}
\caption{Korean conversation dataset between 2 users}
\end{table}%

\begin{table}[ht]
\begin{small}
\begin{center}
\begin{tabularx}{\textwidth}[t]{XX}
\arrayrulecolor{black}\hline
\\
\textbf{\textcolor{black}{Korean ground truth}} & \textbf{\textcolor{black}{Ground truth translated into English}}
\\
\\
\hline
\\
\hspace{0.2em}
\begin{minipage}[t]{\linewidth}
\begin{itemize}
\item schedule.week: 다음주
\item schedule.date: 일요일  
\item schedule.hour: 7시
\item schedule.ampm: 저녁 
\item schedule.location: 스타벅스 
\end{itemize} 
\end{minipage}
\hspace{2em}
&
\hspace{1em}
\begin{minipage}[t]{\linewidth}
\begin{itemize}
\item schedule.week: next week
\item schedule.date: Sunday
\item schedule.hour: 7
\item schedule.ampm: dinner
\item schedule.location: Starbucks\\
\end{itemize} 
\end{minipage}\\
\arrayrulecolor{black}\hline
\end{tabularx}
\end{center}
\end{small}
\hfill{}
\caption{Ground truth of Korean conversation dataset}
\end{table}%

\end{document}